%% file: paper.tex
\documentclass[11pt]{article}
\usepackage{acl2015}
\usepackage{times}
\usepackage{url}
\usepackage{latexsym}

\providecommand{\cite}[1]{\citeauthoryear{#1}}
\usepackage{natbib}
\renewcommand{\cite}{\citep}

\usepackage{url}

\usepackage{subfigure}
\usepackage{times,amsmath,epsfig}
\usepackage{graphicx}
\usepackage{url}
\usepackage{enumitem}
\usepackage{amsfonts}

\usepackage{amsmath}
\usepackage[noend]{algpseudocode}
\usepackage[linesnumbered,ruled]{algorithm2e}
\usepackage{mathtools}
\usepackage{makecell}

\newcommand{\SAVE}[1]{}

\usepackage{mdwlist}  
\usepackage{multirow}  
\usepackage{amsmath} 
\usepackage{relsize} 

\usepackage{float}
\usepackage{flushend}

\usepackage{graphicx}
\makeatletter

\newcommand*\bigcdot@[2]{\mathbin{\vcenter{\hbox{\scalebox{#2}{$\m@th#1\bullet$}}}}}
\makeatother

\usepackage{tikz}

\title{
	Enriching Pre-trained Language Model with Entity Information for Relation Classification
}

\author{\textbf{Shanchan Wu} \\
	Alibaba Group (U.S.) Inc., Sunnyvale, CA   \\
	{\tt shanchan.wu@alibaba-inc.com} \\ 
	\\
	\textbf{Yifan He} \\
	Alibaba Group (U.S.) Inc., Sunnyvale, CA  \\
	{\tt y.he@alibaba-inc.com} \\
	\\
}

\date{}

\begin{document}
\maketitle


\input{abstract}

\input{intro}

\input{related}
\input{methods}

\input{experiments}

\input{conclusion}

\bibliographystyle{acl2015}
\bibliography{mynlp}


\end{document}

%% file: abstract.tex
\begin{abstract}
Relation classification is an important NLP task
to extract relations between entities. The 
state-of-the-art methods for relation classification are primarily
based on Convolutional or Recurrent Neural Networks. 
Recently, the pre-trained BERT model
achieves very successful results in many NLP classification  / sequence labeling tasks. Relation classification differs from
those tasks in that it relies on information of both the
sentence and the two target entities.
In this
paper, we propose a model that both leverages the pre-trained BERT language model and incorporates information from the target entities to tackle the relation classification task. 
%
We locate the target entities and transfer the information through the pre-trained architecture
and incorporate the corresponding encoding of the two entities.
We achieve  significant improvement over the state-of-the-art method on the SemEval-2010 task 8 relational dataset. 
\end{abstract}

%% file: intro.tex
\section{Introduction}

The task of relation classification is to predict semantic relations between pairs of nominals. 
Given a sequence of text (usually a sentence) $s$ and a pair of nominals $e_1$ and $e_2$, the objective
is to identify the relation between $e_1$ and $e_2$ \cite{Hendrickx2010_semeval}. 
It is an important NLP task which is normally
used as an intermediate step in variety of NLP
applications. 
The following example shows the
Component-Whole relation between the nominals
``kitchen'' and ``house'':  ``The [kitchen]$_{e1}$ is the last renovated part of the [house]$_{e1}$.''

Recently, deep neural networks have applied to relation classification 
\cite{Socher_EMNLP_2012,Zeng_coling_2014,Yu_NIPS_Worksho_2014,Nogueira_ACL_2015,Huang_COLING_2016,Joohong_Arxiv_2019}.
These methods usually use  some features
derived from lexical resources such as Word-Net or NLP tools such as dependency parsers and
named entity recognizers (NER).

Language model pre-training has been shown to be effective
for improving many natural language processing
tasks \cite{Dai_NIPS_2015,Peters_arxiv_2017,OpenAI_2018_tech,Ruder_ACL_2018,bert_Jacobv_corr_bert_2018}. 
The pretrained model BERT proposed by \cite{bert_Jacobv_corr_bert_2018} 
has especially significant impact. It has been applied to multiple NLP tasks and obtains new state-of-the-art results on eleven tasks. The tasks that BERT has been applied to are typically modeled as classification problems 
and sequence labeling problems.  
It has also been applied to the
SQuAD question answering \cite{Rajpurkar_squad_2016} problem, in
which the objective is to find the starting point and ending point of an answer span. 

As far as we know, the pretrained BERT model \cite{bert_Jacobv_corr_bert_2018} has not been 
applied to relation classification, which relies 
not only on the information of the whole sentence but also 
on the information of the specific target entities.
In this paper, we apply the pretrained BERT model for relation classification. 
We insert special tokens before and after the target entities 
before feeding the text to BERT for fine-tuning, 
in order to
identify the locations of the two target
entities and transfer the information into the BERT model. 
We then locate the positions of the two target entities in the output embedding
from BERT model. We use their embeddings as well as the sentence encoding (embedding of the special first token in the setting of BERT)
as the input to a multi-layer neural network for classification.
By this way, it captures both the semantics of the sentence and
the two target entities to better fit the relation classification task.

Our contributions are as follows: (1) We put forward an innovative approach
to incorporate entity-level information into the pretrained language model for relation classification. (2) We achieve the new state-of-the-art for the relation classification task.

%% file: related.tex
\section{Related Work}  
\label{related}

There has been some work with deep learning methods for relation
classification, such as 
\cite{Socher_EMNLP_2012,Zeng_coling_2014,Yu_NIPS_Worksho_2014,Nogueira_ACL_2015}

MVRNN model \cite{Socher_EMNLP_2012} applies a recursive neural network (RNN)
to relation classification. They assign a matrix-vector representation to every node
in a parse tree and compute the representation for the complete sentence from bottom up
according to the syntactic structure of the parse tree. 
\cite{Zeng_coling_2014} propose a CNN model by  
incorporating both word embeddings and position features as input.
Then they concatenate lexical features and the output from CNN  
into a single vector and feed them into a softmax layer for prediction.
\cite{Yu_NIPS_Worksho_2014} propose a Factor-based 
Compositional Embedding Model (FCM) by constructing sentence-level and 
substructure embeddings from word embeddings, through dependency trees and named entities.
\cite{Santos_ACL_2015} tackle the 
relation classification task 
by ranking with a convolutional
neural network named CR-CNN. 
Their loss function is based on pairwise ranking. In our work, we take advantage of a pre-trained language model for
the relation classification task, without relying on CNN or RNN
architecutures.
\cite{Huang_COLING_2016} utilize a CNN encoder in conjunction with a sentence representation that weights the words by attention between the target entities and the words in the sentence to perform relation classification.
\cite{Wang-ACL2016_relation} propose a convolutional
neural network architecture with two levels of attention in order
to catch the patterns in heterogeneous
contexts to classify relations.
\cite{Joohong_Arxiv_2019} develop an end-to-end
recurrent neural model which incorporates an
entity-aware attention mechanism with a latent entity
typing for relation classification.  

There are some related work on the relation extraction based on distant supervision,
for example, \cite{Mintz_ACL_2009,Hoffmann_ACL_2011,Lin_ACL_2016,Ji_AAAI_2017,Wu_AAAI_2019}. The difference
between relation classification on regular data and on distantly supervised data is
that the latter may contain a large number of noisy labels. In this paper, we focus on the
regular relation classification problem, without noisy labels. 
 

%% file: methods.tex

\begin{figure*}[!t]
	\centering
	\includegraphics[width=0.74\linewidth]{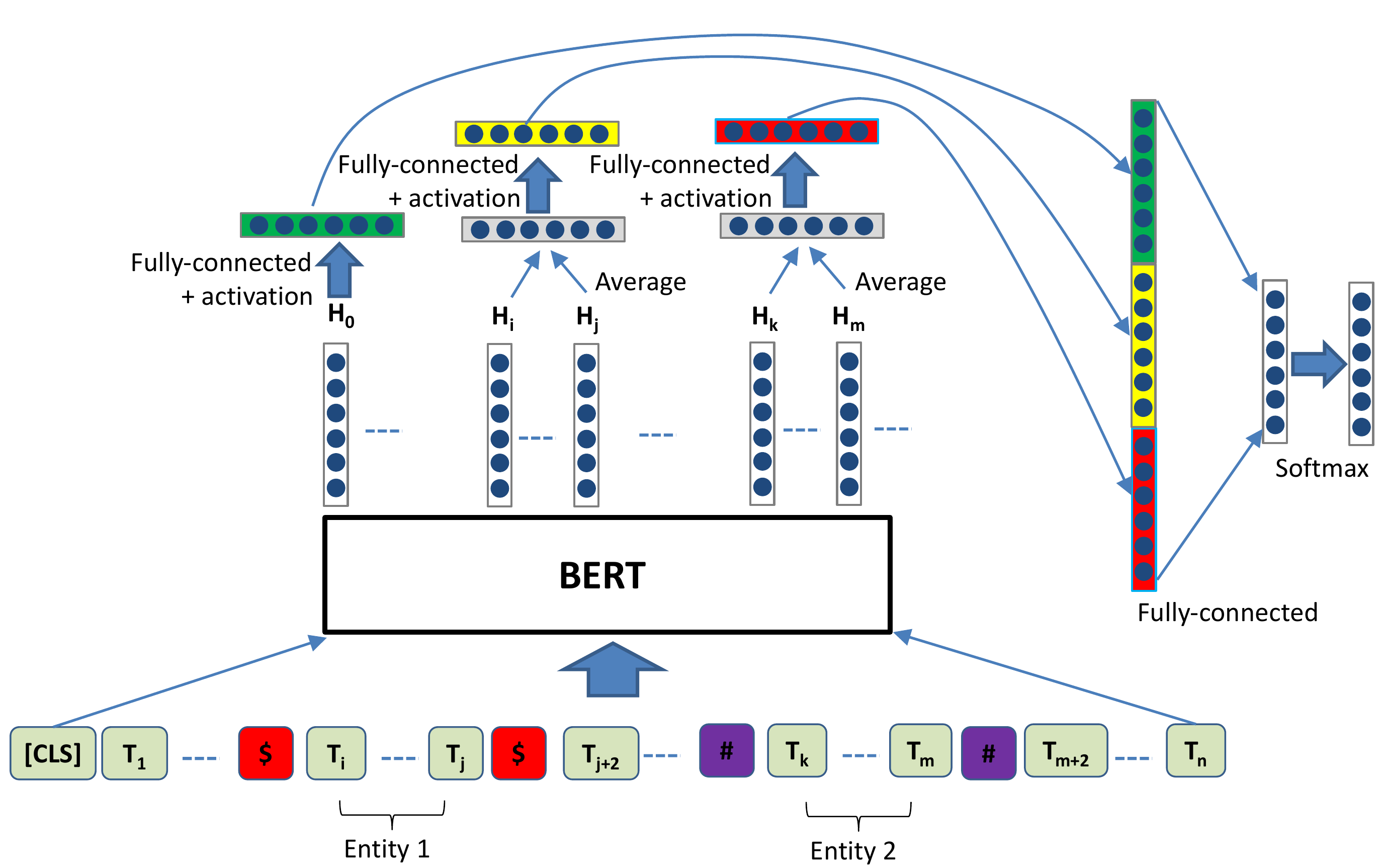}
	\caption{The model architecture. 
	} 
	\label{fig:model_arch}
\end{figure*}


\section{Methodology}  
\label{sec:methods}

\subsection{Pre-trained Model BERT}
The pre-trained BERT model \cite{bert_Jacobv_corr_bert_2018} is a multi-layer bidirectional
Transformer encoder
\cite{Vaswani_NIPS_2017}.

The design of input representation of BERT is to be able to
represent both a single text sentence and a pair of
text sentences in one
token sequence.  The input representation of each token
is constructed by the summation of the corresponding
token, segment and position embeddings.

`[CLS]' is appended to the beginning of each sequence
as the first token of the sequence. The final hidden
state from the Transformer output corresponding to the first token is used as the sentence representation
for classification tasks. In case there are two sentences
in a task, `[SEP]' is used to separate the two sentences. 

BERT pre-trains the model parameters by using a
pre-training objective: the “masked language model” (MLM), which randomly
masks some of the tokens from the input, and set the optimization objective to predict the original vocabulary id of the masked word according to its context. 
Unlike
left-to-right language model pre-training, the
MLM objective can help a state output to utilize both
the left and the right context, which allows a pre-training
system to apply a deep bidirectional Transformer. Besides the masked language model, BERT also
trains a ``next sentence prediction'' task that
jointly pre-trains text-pair representations.



\subsection{Model Architecture}

Figure \ref{fig:model_arch} shows the architecture of our approach.

For a sentence $s$ with two target entities $e_1$ and $e_2$,
to make the BERT module capture the location information of the two entities,
at both the beginning and end of the first entity, we insert
a special token `\$', and at both the beginning and end of the second entity, we insert a special token `\#'. 
We also add `[CLS]' to the beginning of each sentence. 

For example, after insertion of the special separate tokens, for a sentence with target entities ``kitchen'' and ``house'' will become to:

 \textit{``[CLS] The \$ kitchen \$ is the last renovated part of the \# house \# . ''}

Given a sentence $s$ with  entity $e_1$ and $e_2$, suppose its final hidden state output from BERT module is $H$. 
Suppose vectors $H_i$ to $H_j$  are the final hidden state vectors from BERT for
entity $e_1$, and  $H_k$ to $H_m$  are the final hidden state vectors from BERT for
entity $e_2$. We apply the average operation to get a vector representation
for each of the two target entities. Then after an activation operation (i.e. tanh), 
we add a fully connected layer to each of the two vectors, and the output
for $e_1$ and $e_2$ are $H'_1$ and $H'_2$ respectively. This process can be mathematically formalized as Equation (\ref{eq1}).
\begin{equation} \label{eq1}
\begin{split}
&H'_1 =W_1 \left[    tanh \left( \frac{1}{j-i+1} \sum_{t=i}^{j} H_t  \right)      \right] + b_1 \\
&H'_2 =W_2 \left[    tanh \left( \frac{1}{m-k+1} \sum_{t=k}^{m} H_t  \right)      \right]  + b_2\\
\end{split}
\end{equation}

We make $W_1$ and $W_2$, $b_1$ and $b_2$ share the same parameters. In other words,
we set $W_1=W_2$, $b_1=b_2$.
For the final hidden state vector of the first token (i.e. `[CLS]'), we also add an activation operation and a
fully connected layer, which is formally expressed
as:
\begin{equation} \label{eq2}
\begin{split}
&H'_0 = W_0 \left( tanh (H_0)  \right)  + b_0\\
\end{split}
\end{equation}
Matrices $W_0$, $W_1$, $W_2$ have the same dimensions, i.e. 
 $W_0 \in R^{d\times d}$, $W_1 \in R^{d\times d}$, $W_2 \in R^{d\times d}$,
 where $d$ is the hidden state size from BERT.

We concatenate $H'_0$, $H'_1$, $H'_2$ and then add a fully connected layer
and a softmax layer, which can be expressed as following:
\begin{equation} \label{eq3}
\begin{split}
&h''=W_3 \bigg [ concat \left(  H'_0, H'_1, H'_2 \right)  \bigg]  + b_3\\
&p = softmax(h'')
\end{split}
\end{equation}
where $W_3 \in R^{L\times 3d}$ ($L$ is the number of relation types),
and $p$ is the probability output.  In Equations (\ref{eq1}),(\ref{eq2}),(\ref{eq3}),
$b_0, b_1, b_2, b_3$ are bias vectors.

We use cross entropy as the loss function. We apply dropout
before each fully connected layer during training. We call our approach as R-BERT.

%% file: experiments.tex
\section{Experiments } \label{sec:exp}

\subsection{Dataset and Evaluation Metric} \label{sec:dataset}

We use the SemEval-2010 Task 8 dataset in our experiments. 
The dataset contains nine semantic relation types and one
artificial relation type \textit{Other}, which means that the
relation does not belong to any of the nine relation types.
The nine relation types are 
\textit{Cause-Effect,
Component-Whole, Content-Container, Entity-Destination, Entity-Origin, Instrument-Agency,
Member-Collection, Message-Topic} and \textit{Product-Producer}.
The dataset contains
10,717 sentences, with each containing two nominals e1 and e2,
and the corresponding relation type in the sentence. The relation is
directional, which means that Component-Whole(e1, e2) is 
different from Component-Whole(e2, e1). 
The dataset has already been
partitioned into 8,000 training instances and 2,717
test instances. We evaluate our solution by using the
SemEval-2010 Task 8 official scorer script. It computes
the macro-averaged F1-scores for the nine
actual relations (excluding Other) and considers directionality.
%

\subsection{Parameter Settings}

Table  shows the major parameters used in our experiments.

\begin{table}[H]
	\caption{Parameter settings.} \label{tab:para}
	\begin{center}
		\begin{tabular}{ |c|c|c| } 
			\hline
			Batch size   & 16  \\ 
    		Max sentence length & 128 \\
			Adam learning rate  & 2e-5 \\
    		Number of epochs & 5\\ 
			Dropout rate  & 0.1 \\
			\hline
		\end{tabular}
	\end{center}
\end{table}

We add dropout before each add-on layer. 
For the pre-trained BERT model, we use the uncased basic model. 
For the parameters of the pre-trained BERT model, please refer
to \cite{bert_Jacobv_corr_bert_2018} for details. 

\subsection{Comparison with other Methods} \label{sec:compare}

We compare our method, R-BERT, against
results by multiple methods recently published for the SemEval-2010
Task 8 dataset, including SVM, RNN, MVRNN, CNN+Softmax, FCM, CR-CNN, Attention-CNN, Entity Attention Bi-LSTM.
The SVM method by \cite{Rink_Semantic_2010} 
uses a rich feature set in a traditional way,
which was the best result
during the SemEval-2010 task 8 competition. 
Details of all other methods are briefly reviewed in Section \ref{related}.

Table \ref{tab:compare_f1} reports the results. We can see that
R-BERT significantly beats all the baseline methods. The MACRO F1 value
of R-BERT is 89.25, which
is much better than the previous best solution on this dataset.

\begin{table}[h]
	\caption{Comparison with results in the literature.}  
	\label{tab:compare_f1}
	\begin{tabular}{ | c | c | }
		\hline
		\thead{Method} &  \thead{F1} \\
		\hline
		
		\makecell{SVM  \\  \cite{Rink_Semantic_2010}} &  82.2  \\
		
		\hline
		\makecell{RNN \\  \cite{Socher_EMNLP_2012}} &  77.6  \\
		
		\hline
		\makecell{MVRNN \\  \cite{Socher_EMNLP_2012}} &  82.4  \\	
		
		\hline
		\makecell{CNN+Softmax \\  \cite{Zeng_coling_2014}} &  82.7  \\		
		
		\hline
		\makecell{FCM \\  \cite{Yu_NIPS_Worksho_2014}} &  83.0 \\	
		
		\hline
		\makecell{CR-CNN \\  \cite{Santos_ACL_2015}} &   84.1  \\	
		
			\hline
		\makecell{Attention CNN \\  \cite{Huang_COLING_2016}} &   85.9  \\

	\hline
\makecell{Att-Pooling-CNN \\  \cite{Wang-ACL2016_relation}} &   88.0  \\

			\hline
\makecell{Entity Attention Bi-LSTM \\  \cite{Joohong_Arxiv_2019}} &   85.2  \\
		
		\hline
		R-BERT  &   \textbf{89.25}  \\	
		
		\hline
	\end{tabular}
\end{table}

%
%
%
%
%
%
%
%
%
%


\subsection{Ablation Studies} \label{sec:ablation}

\subsubsection{Effect of Model Components}

We have demonstrated the strong empirical results
based on the proposed approach. We further want to
understand the specific contributions by the components besides the 
pre-trained BERT component. For this purpose, we create
three more configurations.  

The first configuration is to discard the special separate tokens (i.e. `\$' and `\#') around
the two entities in the sentence and discard the hidden vector output
of the two entities from concatenating with the hidden vector output 
of the sentence. In other words, we add `[CLS]' at the beginning of the sentence and feed the sentence with the two entities into
the BERT module, and use
the first output vector for classification. We label this method as
\textbf{BERT-NO-SEP-NO-ENT}.

The second configuration is to discard the special separate tokens (i.e. `\$' and `\#') around
the two entities in the sentence, but keep the hidden vector output
of the two entities in concatenation for classification. We label this method as
\textbf{BERT-NO-SEP}.

The third configuration is to discard the hidden vector output
of the two entities from concatenation for classification, but keep the special separate tokens. We label this method as
\textbf{BERT-NO-ENT}.

Table \ref{tab:component_f1} reports the results of the ablation study
with the above three configurations. We observe that the three
methods all perform worse than R-BERT. Of the methods,  BERT-NO-SEP-NO-ENT performs worst, with its F1 8.16 absolute points worse
than R-BERT. This ablation study demonstrates that both the special separate
tokens and the hidden entity vectors make important contributions
to our approach.

 In relation classification, the relation label is dependent on 
both the semantics of the sentence and the two target entities. BERT without
special separate tokens cannot locate the target entities and lose this key 
information. 
The reason why the special separate tokens help to improve the
accuracy is that they identify the locations of the two target entities and 
transfer the information into the BERT model, which make the BERT output
contain the location information of the two entities. On the other hand,
incorporating the output of the target entity vectors further enriches the information and helps to make more accurate prediction.

\begin{table}[H]
	\caption{Comparison of the BERT based methods with different components.}  
	\label{tab:component_f1}
	\begin{tabular}{ | l | c | }
		\hline
		\thead{Method}  & \thead{F1} \\
		\hline
		R-BERT-NO-SEP-NO-ENT  & 81.09  \\	
		\hline
		R-BERT-NO-SEP  & 87.98 \\	
		\hline
		R-BERT-NO-ENT  & 87.99  \\	
		\hline
	    R-BERT  & \textbf{89.25} \\	
		\hline
	\end{tabular}
\end{table}


%
%
%

%% file: conclusion.tex
\section{Conclusions} \label{twitter_sec:conclude}

In this paper, we develop an approach for relation classification by
enriching the pre-trained BERT model
with entity information. We add special separate tokens
to each target entity pair and utilize the sentence vector
as well as target entity representations for classification. We 
conduct experiments on the SemEval-2010 benchmark dataset
and our results significantly outperform the state-of-the-art methods.
One possible future work is to extend the model to apply to 
distant supervision.

%% file: paper.bbl

\begin{thebibliography}{00}


\ifx \showCODEN    \undefined \def \showCODEN     #1{\unskip}     \fi
\ifx \showDOI      \undefined \def \showDOI       #1{#1}\fi
\ifx \showISBNx    \undefined \def \showISBNx     #1{\unskip}     \fi
\ifx \showISBNxiii \undefined \def \showISBNxiii  #1{\unskip}     \fi
\ifx \showISSN     \undefined \def \showISSN      #1{\unskip}     \fi
\ifx \showLCCN     \undefined \def \showLCCN      #1{\unskip}     \fi
\ifx \shownote     \undefined \def \shownote      #1{#1}          \fi
\ifx \showarticletitle \undefined \def \showarticletitle #1{#1}   \fi
\ifx \showURL      \undefined \def \showURL       {\relax}        \fi
\providecommand\bibfield[2]{#2}
\providecommand\bibinfo[2]{#2}
\providecommand\natexlab[1]{#1}
\providecommand\showeprint[2][]{arXiv:#2}

\bibitem[\protect\citeauthoryear{Dai and Le}{Dai and Le}{2015}]%
        {Dai_NIPS_2015}
\bibfield{author}{\bibinfo{person}{Andrew~M Dai} {and} \bibinfo{person}{Quoc~V
  Le}.} \bibinfo{year}{2015}\natexlab{}.
\newblock \showarticletitle{Semi-supervised Sequence Learning}.
\newblock In \bibinfo{booktitle}{{\em Advances in Neural Information Processing
  Systems 28}}. \bibinfo{pages}{3079--3087}.
\newblock


\bibitem[\protect\citeauthoryear{Devlin, Chang, Lee, and Toutanova}{Devlin
  et~al\mbox{.}}{2018}]%
        {bert_Jacobv_corr_bert_2018}
\bibfield{author}{\bibinfo{person}{Jacob Devlin}, \bibinfo{person}{Ming{-}Wei
  Chang}, \bibinfo{person}{Kenton Lee}, {and} \bibinfo{person}{Kristina
  Toutanova}.} \bibinfo{year}{2018}\natexlab{}.
\newblock \showarticletitle{{BERT:} Pre-training of Deep Bidirectional
  Transformers for Language Understanding}.
\newblock \bibinfo{journal}{{\em CoRR\/}}  \bibinfo{volume}{abs/1810.04805}
  (\bibinfo{year}{2018}).
\newblock
\showeprint[arxiv]{1810.04805}


\bibitem[\protect\citeauthoryear{dos Santos, Xiang, and Zhou}{dos Santos
  et~al\mbox{.}}{2015}]%
        {Nogueira_ACL_2015}
\bibfield{author}{\bibinfo{person}{C{\'{\i}}cero~Nogueira dos Santos},
  \bibinfo{person}{Bing Xiang}, {and} \bibinfo{person}{Bowen Zhou}.}
  \bibinfo{year}{2015}\natexlab{}.
\newblock \showarticletitle{Classifying Relations by Ranking with Convolutional
  Neural Networks}. In \bibinfo{booktitle}{{\em Proceedings of the 53rd Annual
  Meeting of the Association for Computational Linguistics and the 7th
  International Joint Conference on Natural Language Processing of the Asian
  Federation of Natural Language Processing, {ACL} 2015}}.
  \bibinfo{pages}{626--634}.
\newblock


\bibitem[\protect\citeauthoryear{Hendrickx, Kim, Kozareva, Nakov, S{\'e}aghdha,
  Pad\'{o}, Pennacchiotti, Romano, and Szpakowicz}{Hendrickx
  et~al\mbox{.}}{2010}]%
        {Hendrickx2010_semeval}
\bibfield{author}{\bibinfo{person}{Iris Hendrickx}, \bibinfo{person}{Su~Nam
  Kim}, \bibinfo{person}{Zornitsa Kozareva}, \bibinfo{person}{Preslav Nakov},
  \bibinfo{person}{Diarmuid~\'{O}. S{\'e}aghdha}, \bibinfo{person}{Sebastian
  Pad\'{o}}, \bibinfo{person}{Marco Pennacchiotti}, \bibinfo{person}{Lorenza
  Romano}, {and} \bibinfo{person}{Stan Szpakowicz}.}
  \bibinfo{year}{2010}\natexlab{}.
\newblock \showarticletitle{SemEval-2010 Task 8: Multi-way Classification of
  Semantic Relations Between Pairs of Nominals}. In \bibinfo{booktitle}{{\em
  Proceedings of the 5th International Workshop on Semantic Evaluation}} {\em
  (\bibinfo{series}{SemEval '10})}. \bibinfo{pages}{33--38}.
\newblock


\bibitem[\protect\citeauthoryear{Hoffmann, Zhang, Ling, Zettlemoyer, and
  Weld}{Hoffmann et~al\mbox{.}}{2011}]%
        {Hoffmann_ACL_2011}
\bibfield{author}{\bibinfo{person}{Raphael Hoffmann}, \bibinfo{person}{Congle
  Zhang}, \bibinfo{person}{Xiao Ling}, \bibinfo{person}{Luke~S. Zettlemoyer},
  {and} \bibinfo{person}{Daniel~S. Weld}.} \bibinfo{year}{2011}\natexlab{}.
\newblock \showarticletitle{Knowledge-Based Weak Supervision for Information
  Extraction of Overlapping Relations}. In \bibinfo{booktitle}{{\em The 49th
  Annual Meeting of the Association for Computational Linguistics, {ACL}
  2011}}. \bibinfo{pages}{541--550}.
\newblock


\bibitem[\protect\citeauthoryear{Ji, Liu, He, and Zhao}{Ji
  et~al\mbox{.}}{2017}]%
        {Ji_AAAI_2017}
\bibfield{author}{\bibinfo{person}{Guoliang Ji}, \bibinfo{person}{Kang Liu},
  \bibinfo{person}{Shizhu He}, {and} \bibinfo{person}{Jun Zhao}.}
  \bibinfo{year}{2017}\natexlab{}.
\newblock \showarticletitle{Distant Supervision for Relation Extraction with
  Sentence-Level Attention and Entity Descriptions}. In
  \bibinfo{booktitle}{{\em Proceedings of the Thirty-First {AAAI} Conference on
  Artificial Intelligence, 2017}}. \bibinfo{pages}{3060--3066}.
\newblock


\bibitem[\protect\citeauthoryear{Lee, Seo, and Choi}{Lee et~al\mbox{.}}{2019}]%
        {Joohong_Arxiv_2019}
\bibfield{author}{\bibinfo{person}{Joohong Lee}, \bibinfo{person}{Sangwoo Seo},
  {and} \bibinfo{person}{Yong~Suk Choi}.} \bibinfo{year}{2019}\natexlab{}.
\newblock \showarticletitle{Semantic Relation Classification via Bidirectional
  {LSTM} Networks with Entity-aware Attention using Latent Entity Typing}.
\newblock \bibinfo{journal}{{\em CoRR\/}} (\bibinfo{year}{2019}).
\newblock


\bibitem[\protect\citeauthoryear{Lin, Shen, Liu, Luan, and Sun}{Lin
  et~al\mbox{.}}{2016}]%
        {Lin_ACL_2016}
\bibfield{author}{\bibinfo{person}{Yankai Lin}, \bibinfo{person}{Shiqi Shen},
  \bibinfo{person}{Zhiyuan Liu}, \bibinfo{person}{Huanbo Luan}, {and}
  \bibinfo{person}{Maosong Sun}.} \bibinfo{year}{2016}\natexlab{}.
\newblock \showarticletitle{Neural Relation Extraction with Selective Attention
  over Instances}. In \bibinfo{booktitle}{{\em Proceedings of the 54th Annual
  Meeting of the Association for Computational Linguistics, {ACL} 2016}}.
\newblock


\bibitem[\protect\citeauthoryear{Mintz, Bills, Snow, and Jurafsky}{Mintz
  et~al\mbox{.}}{2009}]%
        {Mintz_ACL_2009}
\bibfield{author}{\bibinfo{person}{Mike Mintz}, \bibinfo{person}{Steven Bills},
  \bibinfo{person}{Rion Snow}, {and} \bibinfo{person}{Dan Jurafsky}.}
  \bibinfo{year}{2009}\natexlab{}.
\newblock \showarticletitle{Distant Supervision for Relation Extraction Without
  Labeled Data}. In \bibinfo{booktitle}{{\em Proceedings of the Joint
  Conference of the 47th Annual Meeting of the ACL and the 4th International
  Joint Conference on Natural Language Processing of the AFNLP}} {\em
  (\bibinfo{series}{ACL '09})}. \bibinfo{pages}{1003--1011}.
\newblock


\bibitem[\protect\citeauthoryear{Peters, Ammar, Bhagavatula, and Power}{Peters
  et~al\mbox{.}}{2017}]%
        {Peters_arxiv_2017}
\bibfield{author}{\bibinfo{person}{Matthew~E. Peters}, \bibinfo{person}{Waleed
  Ammar}, \bibinfo{person}{Chandra Bhagavatula}, {and} \bibinfo{person}{Russell
  Power}.} \bibinfo{year}{2017}\natexlab{}.
\newblock \showarticletitle{Semi-supervised sequence tagging with bidirectional
  language models}.
\newblock \bibinfo{journal}{{\em CoRR\/}}  \bibinfo{volume}{abs/1705.00108}
  (\bibinfo{year}{2017}).
\newblock
\showeprint[arxiv]{1705.00108}


\bibitem[\protect\citeauthoryear{Radford, Narasimhan, Salimans, and
  Sutskever}{Radford et~al\mbox{.}}{2018}]%
        {OpenAI_2018_tech}
\bibfield{author}{\bibinfo{person}{Alec Radford}, \bibinfo{person}{Karthik
  Narasimhan}, \bibinfo{person}{Tim Salimans}, {and} \bibinfo{person}{Ilya
  Sutskever}.} \bibinfo{year}{2018}\natexlab{}.
\newblock \showarticletitle{Improving language understanding with unsupervised
  learning}.
\newblock \bibinfo{journal}{{\em Technical report, OpenAI\/}}
  (\bibinfo{year}{2018}).
\newblock


\bibitem[\protect\citeauthoryear{Rajpurkar, Zhang, Lopyrev, and
  Liang}{Rajpurkar et~al\mbox{.}}{2016}]%
        {Rajpurkar_squad_2016}
\bibfield{author}{\bibinfo{person}{Pranav Rajpurkar}, \bibinfo{person}{Jian
  Zhang}, \bibinfo{person}{Konstantin Lopyrev}, {and} \bibinfo{person}{Percy
  Liang}.} \bibinfo{year}{2016}\natexlab{}.
\newblock \showarticletitle{SQuAD: 100,000+ Questions for Machine Comprehension
  of Text}. In \bibinfo{booktitle}{{\em Proceedings of the 2016 Conference on
  Empirical Methods in Natural Language Processing}}.
  \bibinfo{publisher}{Association for Computational Linguistics},
  \bibinfo{pages}{2383--2392}.
\newblock


\bibitem[\protect\citeauthoryear{Rink and Harabagiu}{Rink and
  Harabagiu}{2010}]%
        {Rink_Semantic_2010}
\bibfield{author}{\bibinfo{person}{Bryan Rink} {and} \bibinfo{person}{Sanda
  Harabagiu}.} \bibinfo{year}{2010}\natexlab{}.
\newblock \showarticletitle{Utd: Classifying semantic relations by combining
  lexical and semantic resources}. In \bibinfo{booktitle}{{\em Proceedings of
  the 5th International Workshop on Semantic Evaluation}}.
  \bibinfo{pages}{256--259}.
\newblock


\bibitem[\protect\citeauthoryear{Ruder and Howard}{Ruder and Howard}{2018}]%
        {Ruder_ACL_2018}
\bibfield{author}{\bibinfo{person}{Sebastian Ruder} {and}
  \bibinfo{person}{Jeremy Howard}.} \bibinfo{year}{2018}\natexlab{}.
\newblock \showarticletitle{Universal Language Model Fine-tuning for Text
  Classification}. In \bibinfo{booktitle}{{\em Proceedings of the 56th Annual
  Meeting of the Association for Computational Linguistics, {ACL} 2018,
  Melbourne, Australia, July 15-20, 2018, Volume 1: Long Papers}}.
  \bibinfo{pages}{328--339}.
\newblock


\bibitem[\protect\citeauthoryear{Santos, Xiang, and Zhou}{Santos
  et~al\mbox{.}}{2015}]%
        {Santos_ACL_2015}
\bibfield{author}{\bibinfo{person}{Cicero Nogueira~Dos Santos},
  \bibinfo{person}{Bing Xiang}, {and} \bibinfo{person}{Bowen Zhou}.}
  \bibinfo{year}{2015}\natexlab{}.
\newblock \showarticletitle{Classifying Relations by Ranking with Convolutional
  Neural Networks}. In \bibinfo{booktitle}{{\em Proceedings of the 53rd Annual
  Meeting of the Association for Computational Linguistics and the 7th
  International Joint Conference on Natural Language Processing of the Asian
  Federation of Natural Language Processing, (ACL) 2015}}.
\newblock


\bibitem[\protect\citeauthoryear{Shen and Huang}{Shen and Huang}{2016}]%
        {Huang_COLING_2016}
\bibfield{author}{\bibinfo{person}{Yatian Shen} {and} \bibinfo{person}{Xuanjing
  Huang}.} \bibinfo{year}{2016}\natexlab{}.
\newblock \showarticletitle{Attention-based Convolutional Neural Network for
  Semantic Relation Extraction}. In \bibinfo{booktitle}{{\em Proceedings of
  COLING 2016, the 26th International Conference on Computational Linguistics:
  Technical Papers}}. \bibinfo{pages}{2526--2536}.
\newblock


\bibitem[\protect\citeauthoryear{Socher, Huval, Manning, and Ng}{Socher
  et~al\mbox{.}}{2012}]%
        {Socher_EMNLP_2012}
\bibfield{author}{\bibinfo{person}{Richard Socher}, \bibinfo{person}{Brody
  Huval}, \bibinfo{person}{Christopher~D. Manning}, {and}
  \bibinfo{person}{Andrew~Y. Ng}.} \bibinfo{year}{2012}\natexlab{}.
\newblock \showarticletitle{Semantic Compositionality Through Recursive
  Matrix-vector Spaces}. In \bibinfo{booktitle}{{\em Proceedings of the 2012
  Joint Conference on Empirical Methods in Natural Language Processing and
  Computational Natural Language Learning}} {\em (\bibinfo{series}{EMNLP-CoNLL
  '12})}. \bibinfo{address}{Stroudsburg, PA, USA}, \bibinfo{pages}{1201--1211}.
\newblock


\bibitem[\protect\citeauthoryear{Vaswani, Shazeer, Parmar, Uszkoreit, Jones,
  Gomez, Kaiser, and Polosukhin}{Vaswani et~al\mbox{.}}{2017}]%
        {Vaswani_NIPS_2017}
\bibfield{author}{\bibinfo{person}{Ashish Vaswani}, \bibinfo{person}{Noam
  Shazeer}, \bibinfo{person}{Niki Parmar}, \bibinfo{person}{Jakob Uszkoreit},
  \bibinfo{person}{Llion Jones}, \bibinfo{person}{Aidan~N Gomez},
  \bibinfo{person}{\L~ukasz Kaiser}, {and} \bibinfo{person}{Illia Polosukhin}.}
  \bibinfo{year}{2017}\natexlab{}.
\newblock \showarticletitle{Attention is All you Need}.
\newblock In \bibinfo{booktitle}{{\em Advances in Neural Information Processing
  Systems 30}}, \bibfield{editor}{\bibinfo{person}{I.~Guyon},
  \bibinfo{person}{U.~V. Luxburg}, \bibinfo{person}{S.~Bengio},
  \bibinfo{person}{H.~Wallach}, \bibinfo{person}{R.~Fergus},
  \bibinfo{person}{S.~Vishwanathan}, {and} \bibinfo{person}{R.~Garnett}}
  (Eds.). \bibinfo{pages}{5998--6008}.
\newblock


\bibitem[\protect\citeauthoryear{Wang, Cao, de~Melo, and Liu}{Wang
  et~al\mbox{.}}{2016}]%
        {Wang-ACL2016_relation}
\bibfield{author}{\bibinfo{person}{Linlin Wang}, \bibinfo{person}{Zhu Cao},
  \bibinfo{person}{Gerard de Melo}, {and} \bibinfo{person}{Zhiyuan Liu}.}
  \bibinfo{year}{2016}\natexlab{}.
\newblock \showarticletitle{Relation Classification via Multi-Level Attention
  {CNN}s}. In \bibinfo{booktitle}{{\em Proceedings of the 54th Annual Meeting
  of the Association for Computational Linguistics}}.
\newblock


\bibitem[\protect\citeauthoryear{Wu, Kai, and Zhang}{Wu et~al\mbox{.}}{2019}]%
        {Wu_AAAI_2019}
\bibfield{author}{\bibinfo{person}{Shanchan Wu}, \bibinfo{person}{Fan Kai},
  {and} \bibinfo{person}{Qiong Zhang}.} \bibinfo{year}{2019}\natexlab{}.
\newblock \showarticletitle{Improving Distantly Supervised Relation Extraction
  with Neural Noise Converter and Conditional Optimal Selector}. In
  \bibinfo{booktitle}{{\em Proceedings of the Thirty-Third {AAAI} Conference on
  Artificial Intelligence, 2019}}.
\newblock


\bibitem[\protect\citeauthoryear{Yu, Gormley, and Dredze}{Yu
  et~al\mbox{.}}{2014}]%
        {Yu_NIPS_Worksho_2014}
\bibfield{author}{\bibinfo{person}{Mo Yu}, \bibinfo{person}{Matthew~R.
  Gormley}, {and} \bibinfo{person}{Mark Dredze}.}
  \bibinfo{year}{2014}\natexlab{}.
\newblock \showarticletitle{Factor-based compositional embedding models}. In
  \bibinfo{booktitle}{{\em In NIPS Workshop on Learning Semantics}}.
\newblock


\bibitem[\protect\citeauthoryear{Zeng, Liu, Lai, Zhou, and Zhao}{Zeng
  et~al\mbox{.}}{2014}]%
        {Zeng_coling_2014}
\bibfield{author}{\bibinfo{person}{Daojian Zeng}, \bibinfo{person}{Kang Liu},
  \bibinfo{person}{Siwei Lai}, \bibinfo{person}{Guangyou Zhou}, {and}
  \bibinfo{person}{Jun Zhao}.} \bibinfo{year}{2014}\natexlab{}.
\newblock \showarticletitle{Relation Classification via Convolutional Deep
  Neural Network}. In \bibinfo{booktitle}{{\em {COLING} 2014, 25th
  International Conference on Computational Linguistics, Proceedings of the
  Conference: Technical Papers, 2014}}. \bibinfo{pages}{2335--2344}.
\newblock


\end{thebibliography}
